\documentclass[review]{elsarticle}

\usepackage[]{caption2}

\usepackage{amsmath}
\usepackage{subfigure}
\usepackage{url}
\usepackage{lineno,hyperref}
\usepackage{amssymb}
\usepackage{booktabs}
\usepackage{multirow}
\usepackage{color}
\usepackage{multicol}
\usepackage{amsfonts}
\usepackage{textcomp}
\usepackage{changepage}
\usepackage{algorithm} 
\usepackage{algorithmic}
\usepackage{geometry}
\usepackage{bm}
\usepackage{setspace}
\modulolinenumbers[5]
\journal{Journal of \LaTeX\ Templates}
\usepackage{array}
\newcommand{\PreserveBackslash}[1]{\let\temp=\\#1\let\\=\temp}
\newcolumntype{C}[1]{>{\PreserveBackslash\centering}p{#1}}
\newcolumntype{R}[1]{>{\PreserveBackslash\raggedleft}p{#1}}
\newcolumntype{L}[1]{>{\PreserveBackslash\raggedright}p{#1}}

\biboptions{sort&compress}

\begin{document}

\begin{frontmatter}

\title{Hyperspectral Remote Sensing Benchmark Database for Oil Spill Detection with an Isolation Forest-Guided Unsupervised Detector}


\author[mymainaddress]{Puhong Duan}
\author[mymainaddress]{Xudong Kang}
\ead{xudong\_kang@163.com}
\author[myfourthaddress]{Pedram Ghamisi}


\address[mymainaddress]{College of Electrical and Information Engineering, Hunan University, 410082 Changsha, China}
\address[myfourthaddress]{Helmholtz-Zentrum Dresden-Rossendorf, Helmholtz Institute Freiberg for Resource Technology, Freiberg 09599, Germany}
\begin{abstract}
Oil spill detection has attracted increasing attention in recent years since marine oil spill accidents severely affect environments, natural resources, and the lives of coastal inhabitants. Hyperspectral remote sensing images provide rich spectral information which is beneficial for the monitoring of oil spills in complex ocean scenarios. However, most of the existing approaches are based on supervised and semi-supervised frameworks to detect oil spills from hyperspectral images (HSIs), which require a huge amount of effort to annotate a certain number of high-quality training sets. In this study, we make the first attempt to develop an unsupervised oil spill detection method based on isolation forest for HSIs. First, considering that the noise level varies among different bands, a noise variance estimation method is exploited to evaluate the noise level of different bands, and the bands corrupted by severe noise are removed. Second, the kernel principal component analysis (KPCA) is employed to reduce the high dimensionality of the HSIs. Then, the probability of each pixel belonging to one of the classes of seawater and oil spills is estimated with the isolation forest, and a set of pseudo labeled training samples is automatically produced using the clustering algorithm on the detected probability. Finally, an initial detection map can be obtained by performing the support vector machine (SVM) on the dimension-reduced data, and then, the initial detection result is further optimized with the extended random walker (ERW) model so as to improve the detection accuracy of oil spills. To evaluate the performance of the proposed approach, we create a database, containing eighteen hyperspectral images with oil spills captured over Gulf of Mexico in 2010, entitled airborne hyperspectral oil spill data (HOSD). The obtained results demonstrate that the detection accuracies achieved by the proposed method have been increased by 21.14 percentage points and 33.29 percentage points in terms of area under curve (AUC) and detection precision (DP) in comparison with other state-of-the-art detection approaches. We will make HOSD and our developed library for oil spill detection publicly available at \url{https://github.com/PuhongDuan/HOSD} to further promote this research topic.
\end{abstract}


\end{frontmatter}


\section{Introduction}
\label{sec:intro}
In the wake of marine oil exploration and transportation, the accidents of oil spills have occurred  frequently around the world, which leads to the severe pollution of the marine environment and the huge damage of coastal species \citep{ALPERS2017133,BREKKE20051,KONIK201637,Boc,PELTA2019101901,PELTA2019145}. On April 20, 2010, the explosion of Deepwater Horizon oil drilling platform led to a severe leakage. Million barrels of oil polluted the Gulf of Mexico with the area of about 10,000 square kilometers \citep{jason,YingLi}. Due to this accident, the marine ecosystems, such as fish and seabirds, have been seriously destroyed. On June 4, 2011, the Penglai 19-3 oilfield in Bohai Bay, Northeast China, witnessed a serious oil spill incident, which caused the leak of more than 7 thousand tons of oil into the sea \citep{junfang}. The polluted area was almost 6200 square kilometers. If the oil would not be timely monitored after the leakage, the oil slick would be washed onto the coast by the sea waves. This situation would pose a huge threat to coastal aquaculture fishery resources and human health. Therefore, it is of great importance to effectively detect oil spills on the sea surface to monitor the distribution, impact, and volume of oil spills \citep{BHANGALE201728}.\par
Remote sensing techniques make it possible to monitor the spatial distribution of oil spills in remote and inaccessible areas on a large size \citep{LBing}. In addition, the speed and direction of the movement of oil spills are also estimated by using multi-temporal data and drift prediction models, which play an important role in facilitating clean-up tasks and warning systems \citep{Karantzalos2008}. In the past decades, researchers have explored the capability of remote sensing data in detecting and monitoring the oil spill regions. The earliest application of remote sensing for oil spill detection was based on airborne visible (VIS) and infrared (IR) data \citep{FINGAS20149}. However, this type of data suffers from diverse disadvantages, such as the low separability between the oil spills and the surrounding objects \citep{6488890}.\par
Different from VIS and IR data, active microwave satellite sensors, particularly synthetic aperture radar (SAR), have been widely investigated for oil spill detection due to their day-and-night and all-weather imaging advantages \citep{SHU20102026,LI20101590,BREKKE20141,LATINI201626,RadarMM,RadarMM02}. The main principle of SAR oil spill detection is that oil film has an inhibitory effect on the capillary waves, and hence it reduces the backscatter of the ocean surface, resulting in the appearance of dark spots in SAR images compared to the surrounding spill-free regions. For example, a generalized-likelihood ratio test-based edge detector was proposed based on SAR data to detect dark spots on the ocean surface \citep{6563113,1468095}. A polarimetric analysis framework was developed to analyze and compare the physical relationship between fully polarimetric (FP) and compact-polarimetric (CP) SAR architectures for ocean oil slick estimation. Experiment analysis shown that FP and CP SAR led to a similar performance while the $\pi/4$ mode performs the best in detecting oil slicks \citep{7501602,4689341,6463451}. These publications have demonstrated that SAR data can be effectively applied for detecting oil spill areas on the ocean surface. However, a principal challenge of oil spill detection with SAR data is to identify the oil spill regions from other natural phenomena, since the darkened patches in the SAR image may be caused by grease ice, wind sheltering by land, internal waves, etc. \citep{8357531}. Furthermore, the other limitations of the SAR data are its high cost, low revisit frequencies, and relatively small swath widths. These limitations call for the development of the multi-platform SAR and other optical sensors, such as multispectral and hyperspectral sensors.\par
In addition to the SAR-based approaches, many studies have examined the feasibility of the multispectral image (MSI) for oil spill detection with high-level reliability. For instance, the Moderate Resolution Imaging Spectroradiometer (MODIS) instrument has been demonstrated to be effective in identifying oil spill regions due to relatively high spectral and temporal resolutions \citep{Zhao14,6069530,rs70101112,OD,rs9020128,rs11232762,7835233}. Besides, Landsat data with a higher spatial resolution have also been used to detect the oil spill regions by using the spectral and spatial information \citep{la1,HESE2009130,SUN2015632,la2}. Most of these oil spill detection methods are very analogous to the semantic segmentation task in computer vision, which relies on the labeled samples to learn a supervised classifier. However, MODIS data with 250 m resolution and Landsat with 30 m resolution may not meet the demand for fine oil spill identification of small regions, since the spatial resolution of the MSI data is greatly larger than the size of oil spill regions.\par
Hyperspectral sensors mounted on aircraft, an emerging technique in the remote sensing field, can provide both rich spectral and spatial resolutions. Such sensors can record hundreds of narrow spectral channels from the visible to the infrared wavelength. In recent years, this technique has been extensively applied in all kinds of aspects, such as land cover classification \citep{8474403,7882742,8351989}, retrieval of water constituents \citep{BRANDO2009755,DUAN2020359}, and mineral mapping \citep{9098893,8758206,rs12182903}. Two types of oil emulsions, e.g., water in oil (WO) and oil in water (OW), were investigated with hyperspectral data, which have illustrated that feature bands at 1677 nm and 839 nm can be used to distinguish the WO and OW emulsions \citep{LU2019111183,LU2020111778}. Besides, machine learning-based oil spill detection techniques with this type of data have been also developed, which was to find the spectral signature of the oil from the complex ocean environment, which could greatly decrease the false alarm rate of the oil spill regions \citep{LEIFER2012185,KOKALY2013210,hsi1,hsi2,8529276}. For example, in \citep{7835176,PETERSON2015222,Liu}, spectral shape matching methods were developed to achieve oil spill detection. Given a reference spectrum selected from the original image, the oil spill regions can be detected by matching the spectral curve shape of the selected spectrum and one of the observed spectrum. In \citep{DONGMEI2015133,junfang,ijgi8040181,SHI2018251}, spatial and spectral feature extraction methods were introduced for oil spill detection, which can decrease the influence of the shadows and image noise, as well as increase the discrimination between the oil spill and the seawater. These approaches have been confirmed to be effective in identifying oil spill regions with a certain number of training set.
\par
Nevertheless, there are still several challenges in achieving accurate oil spill detection with hyperspectral images, which are presented as follows:
\begin{itemize}
 \item{}\textit{The lack of well-annotated public oil spill dataset}: Up to now, many oil spill detection approaches have been proposed in the remote sensing community, including machine learning algorithms and deep learning models. However, researchers often adopt a few HSIs to verify the effectiveness of the  method, which may decrease the generalization ability of those approaches. Furthermore, those datasets are not available for other researchers, which cannot effectively promote the development of marine oil spill monitoring.
 \item{} \textit{The requirement of the available training samples}: Current hyperspectral oil spill detection approaches are based on supervised or semi-supervised learning manners, and thus, they highly rely on the availability of sufficient training samples. However, the collection of the high-quality training set is time-consuming and expensive because of the spectral variability caused by sun glint intensity, oil thickness, and oil emulsification stage, which leads to the fact that they cannot work in an unsupervised manner.
 \item{} \textit{The negative effect of image noise}: Due to the imaging environment such as low lighting, shadow, and weather, the captured spectral bands in hyperspectral images are usually corrupted by different degrees of noise, and the spectral bands with serious noise have a negative effect on the accuracy of the oil spill detection. In order to reduce the involvement of domain experts for the automatic detection of oil spill regions, the detection methods should have the capability to automatically decrease the interference of the severe noisy channels.
\end{itemize}\par
In order to overcome these issues, this paper proposes an unsupervised oil spill detection method based on isolation forest, in which the original HSI with severe noisy bands can be directly used as input. First, to alleviate the negative effect of the noisy bands, a Gaussian statistical-based method is developed to automatically remove the bands corrupted by serious noise. Then, the isolation forest method is performed on the dimensionality-reduced data so as to generate the pseudo labeled training samples. Next, the SVM classifier is trained using these pseudo labeled samples, which have been produced automatically, to obtain an initial detection map. Finally, in order to further refine the detection result, the ERW is adopted as the postprocessing step of the initial detection map, in which the interactions among adjacent pixels are fully exploited. Experiments performed on the HOSD database validate that the proposed method obtains promising detection performance under different scenes with respect to other state-of-the-art approaches in terms of both visual map and objective quality.\par
Compared to previous studies, the novel contributions of the proposed oil spill detection method are concluded as follows: \par
\begin{itemize}
\item{} We construct a novel hyperspectral remote sensing database, named \textit{HOSD}, for oil spill detection, which is a publicly available benchmark dataset. To the best of our knowledge, this is the first public dataset for oil spill detection. Moreover, the dataset is of the largest scale on the total number of images. The building of this dataset will enable the research community to advance the state-of-the-art algorithms for oil spill detection.
\item{} Different from previous supervised oil spill detection methods which require the users to annotate the oil spill and seawater samples, we propose an unsupervised oil spill detection method based on isolation forest for the first time. Both the spectral and spatial information of oil spill regions are considered by the proposed method to automatically generate training samples for the spectral classifier.
\item{} The Gaussian statistical method is explored to estimate the noise level of each spectral band, and the bands contaminated with serious noise are automatically removed. Experimental results demonstrate that noisy band removal is beneficial for oil spill detection.
\end{itemize}

The rest of this work is summarized as follows. Section \ref{sec:proposed} gives the details of the proposed method. Section \ref{sec:data} depicts the study area and the constructed dataset. The comprehensive experiments performed on the constructed dataset are presented in Section \ref{sec:exp}, followed by an analysis in Section \ref{sec:dis}. Finally, we summarize this study in Section \ref{sec:con}.\par
\begin{figure*}[!tp]
\centering
\includegraphics[scale=0.43]{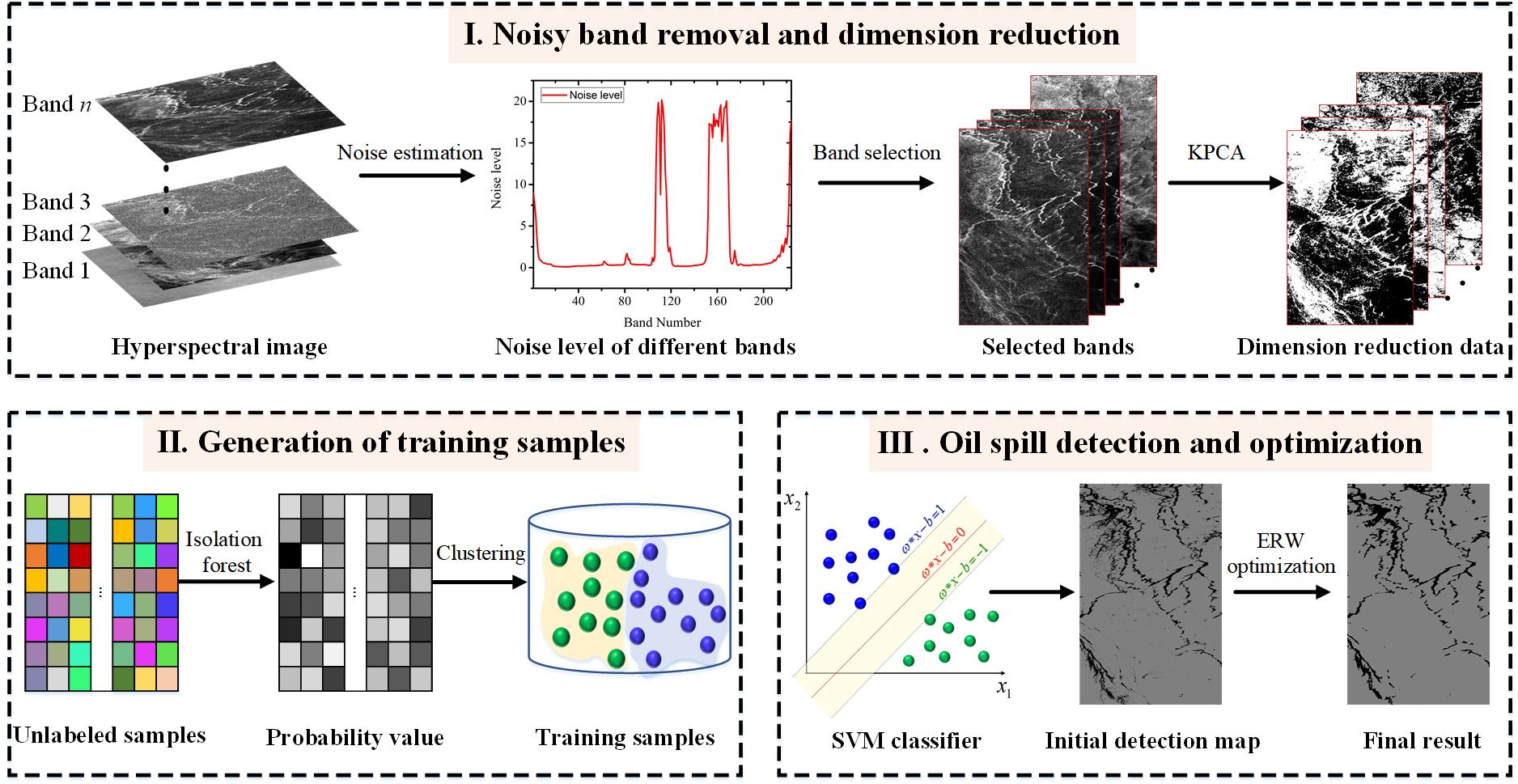}
\caption{The flowchart of the proposed unsupervised oil spill detection method.}
\label{schematic}
\end{figure*}

\section{Methodology}
\label{sec:proposed}
To automatically and efficiently detect oil spills using hyperspectral images, we develop an unsupervised method based on isolation forest. As shown in Fig. \ref{schematic}, the proposed method mainly contains three steps. First, the noise variance of each band is estimated with the Gaussian statistical method, and the important channels are selected from the hundreds of spectral bands according to the noise level. Then, the number of selected bands is further decreased using KPCA. Second, the isolation forest is employed to detect the probability of each pixel belonging to the oil film and the seawater, and the $k$-means method is used to construct the positive and negative samples. Finally, the training set is fed into the SVM classifier to yield the initial detection map, and the ERW algorithm is utilized to optimize the initial detection map by integrating the spatial information so as to produce the final result. The detailed steps of the proposed approach are depicted in the following sections.

\subsection{Noisy band removal and dimension reduction}
Due to the sensor instability, calibration error, and photon effects, the captured hyperspectral images often suffer from noise contamination, which seriously degrades the quality of hyperspectral images and has a negative effect on image interpretation, such as object identification, image classification, and so on \citep{7167714,KANG2020130,HONG202012}. In the hyperspectral image classification community, a common strategy is to manually remove the bands corrupted by serious noise, and then the remaining bands are used for the subsequent tasks \citep{MPCNN,MEPOs,Rasti2020}. However, this band selection manner is time-consuming because of the high spectral dimension of hyperspectral data. To alleviate the issue, in this paper, we develop an automatic band selection method based on the Gaussian statistical model to eliminate the severe noise bands. The band importance is first evaluated, and then the spectral bands with low noise variance are chosen based on an adaptive thresholding method. Specifically, the Gaussian statistical model \citep{IMMERKAER1996300} is exploited to measure the noise level of different spectral bands. The noise variance of each spectral band is calculated by
\begin{equation}
\label{noise}
{\sigma _n} = \sqrt {\frac{\pi }{2}} \frac{1}{{6(I_W - 2)(I_H - 2)}}\sum\limits_{i,j} {\left| {\mathbf{I}_n(i,j) * \mathbf{M}} \right|}, n=1,2,..,I_N
\end{equation}
where $I_W$ and $I_H$ stand for the spatial dimensions. $I_N$ is the total number of spectral channels. $\mathbf{M}$ is the image mask which is used to estimate the amount of noise in local region. Here, the Laplacian operator is considered as the image mask since it is sensitive to image noise.
\begin{equation}
\mathbf{M} = \left[ {\begin{array}{*{20}{c}}
1&{ - 2}&1\\
{ - 2}&4&{ - 2}\\
1&{ - 2}&1
\end{array}} \right]
\end{equation}
Accordingly, the noise level of different bands is calculated by Eq. (\ref{noise}), which measures the importance of spectral bands with respect to oil spill detection. Next, the bands contaminated with serious noise can be easily determined and removed by using a threshold. The spectral bands with noise variance less than the threshold are preserved for the subsequent identification.
\begin{equation}
\label{eq:denoising}
{\mathbf{S}_n} = \left\{ \begin{array}{l}
{\mathbf{I}_n},\;\;if\;{\sigma _n} < \frac{1}{{2I_N}}\sum\limits_n {{\sigma _n}} \\
\phi ,\;\;Otherwise
\end{array} \right.
\end{equation}
Although the selected bands are more useful for identifying oil spill regions, the spectral dimension of the selected channels is still high. In addition, the spectral separability of pixels between seawater and oil film in the obtained data is relatively low. In order to improve discriminative capabilities for identifying oil spill regions and decrease the computing cost spent in the following operations, KPCA \citep{KPCA} is utilized to fuse the spectral dimension of the selected data. Specifically, the selected data $\mathbf{S}=\{\mathbf{s}_1,\mathbf{s}_2,...,\mathbf{s}_m\}\in {\cal{R}}^{I_N}$ is first projected into high-dimensional feature space $\cal{H}$ by using a kernel function $\Phi :\;{{\cal{R}}^{I_N}} \to \;\cal{H}$, where the radial basis function kernel is adopted. Then, the kernel principal components can be obtained:
\begin{equation}
\label{eq:kpca}
\mathbf{K}\bm{\alpha}  = \lambda \bm{\alpha} \quad s.t.\quad {\left\| \bm{\alpha}  \right\|_2} = \frac{1}{\lambda }
\end{equation}
where $\mathbf{K}$ denotes the Gram matrix ${{\Phi} ^T}(\mathbf{S}){\Phi} (\mathbf{S})$. In this work, the selected bands $\mathbf{S}\in {\cal{R}}^{I_N}$ are projected into $\mathbf{Y}\in{\cal{R}}^{D} $. In this work, we refer the KPCA with $\mathbf{Y}=\tt{KPCA}(\mathbf{S}, D)$, where $D$ is the number of bands.\par
\subsection{Generation of training samples}\par
\begin{figure}[!tp]
\footnotesize
\setlength{\abovecaptionskip}{0.cm}
\setlength{\belowcaptionskip}{-0.cm}
\centering
\centerline{\subfigure[]{\includegraphics[width=1.2in,height=1.5in]{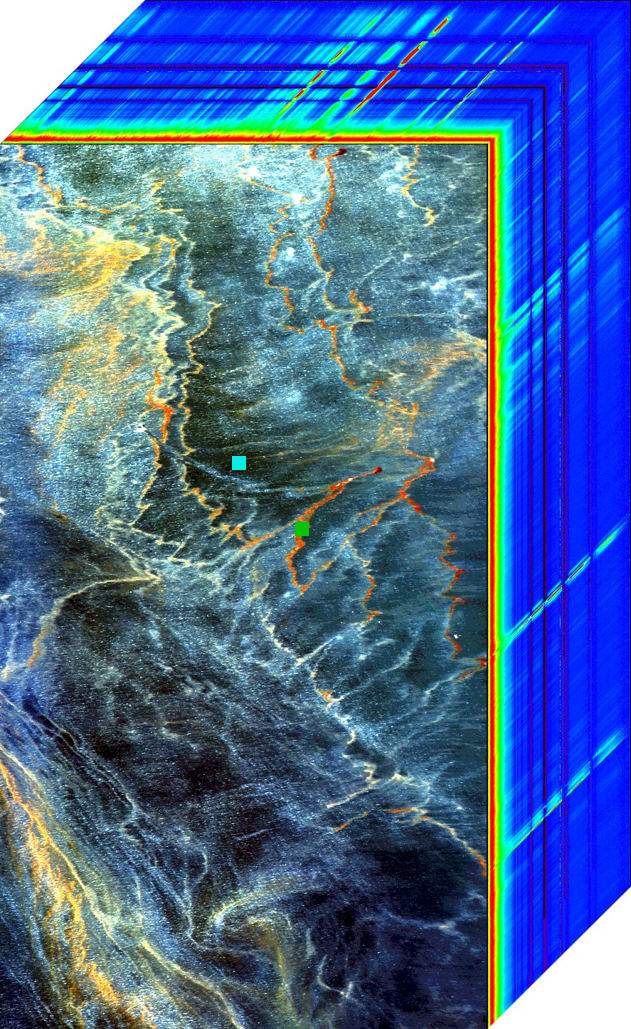}}
			\subfigure[]{\includegraphics[width=2in,height=1.5in]{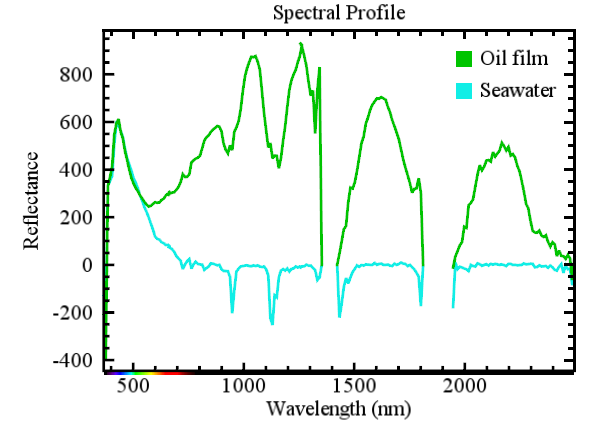}}}
\caption{The spectral curve of different objects. (a) Hyperspectral oil spill image. (b) Spectral curve.}
\label{fig:curve}
\end{figure}

\sloppy{}
Supervised detection approaches usually require a certain number of training samples for different classes. However, human labeling is time-consuming and an expensive task \citep{TONG2020111322}. To alleviate this issue, we propose an automatic scheme for producing the training set instead of manual interpretation. By comparing the spectral curves of oil film and seawater, it is found that the spectral value of oil film is different from the surrounding seawater. Therefore, the spectral pixels belonging to oil film are easier to be distinguished than the seawater. Based on this observation, the isolation forest (iForest) is adopted to obtain the probability of each pixel. \par
iForest is first proposed by Liu \emph{et al.} for identifying outliers in machine learning \citep{Liu2012RF}, which has been widely used in image processing \citep{8833502,8960541}. The iForest is based on an assumption that the outliers are often more susceptible to be isolated in a given data, and thus the probability value of the same object tends to be the same. In oil spill detection, due to the spectral difference between the oil film and the seawater (see Fig. \ref{fig:curve}), the oil film is also easily isolated. Accordingly, the probability value belonging to oil film or seawater is also larger. Specifically, the automatic generation of training samples is described in detail below:\par

1) {$Isolation$ $forest$ $construction$}: The obtained dimension-reduced data $\mathbf{Y}$ is used to construct the isolation forest. $\mathbf{Y}$ is transformed into two dimensional matrix $\mathbf{\hat{Y}}\in \mathbb{R}^{D\times N}$, where $D$ and $N$ denote the total number of bands and pixels, respectively. In order to construct the isolation forest, first, it is necessary to randomly choose $K$ pixels from $\mathbf{\hat{Y}}$, and divide the selected pixels into two child nodes, i.e., left node and right node. The pixel value smaller than the split value is divided into the left node, and the pixel value larger than or equal to the split value is divided into the right node. Here, the split value is stochastically chosen between the minimum value and maximum value of $\mathbf{\hat{Y}}$. Then, the child nodes are recursively performed until one of the following condition is met: (1) the quantity of pixels in each child node is 1; (2) the tree reaches the maximum height $H_{max}$; (3) the pixels in each child node are the same. In this work, $H_{max}=log_2K$ is regarded as the default parameter. Finally, the construction of isolation tree is repeated $T$ times to obtain the isolation forest, where $T$ is a free parameter.\par
2) $Probability$ $calculation$ $of$ $each$ $pixel$: In this step, the constructed isolation forest is used to estimate the probability value of each pixel. In more detail, for each isolation tree, the path length $h(x)$ of a pixel $x$ is the quantity of edges that the pixel passes through in an isolation tree from the root node to the terminating node. Since the oil film pixels are always different from the surrounding seawater, and thus they are easily isolated to the terminating node compared to seawater pixels. In this situation, the oil film pixels incline to have shorter path lengths in the isolation trees, and the path length is considered as a measure to obtain probability value of each pixel. It should be mentioned that the path length of each pixel in different isolation trees varies. Therefore, the final path length is produced by calculating the average path length of all isolation trees. Assume that the constructed iForest has $T$ isolation trees, and $h_i(x)$ is the path length of a test pixel $x\in\mathbf{\hat{Y}}$ in the $i$th isolation tree, the average path length $A(h(x))$ on $T$ isolation trees can be obtained:
\begin{equation}
A(h(x)) = \frac{1}{T}\sum\limits_{i = 1}^T {{h_i}(x)}
\end{equation}
Accordingly, the probability value $p\in(0,1]$ of the test pixel $x$ is calculated:
\begin{equation}
\label{eq:prob}
p(x) = {2^{ - \frac{{A(h(x))}}{{c(m)}}}}
\end{equation}
where $c(m)=2H(m-1)-\frac{2(m-1)}{m}$. $H(m)$ denotes the harmonic number estimated by $ln(m)+0.5772156649$ (the Euler's constant). After performing the same steps above to each pixel of $\mathbf{\hat{Y}}$, the probability map $\mathbf{p}$ can be obtained.\par
3) $Clustering$: Since the probability of the same object, i.e., oil film or seawater, tends to be the same, the label of each pixel can be decided by using a clustering method, where the number of clusters is 2, i.e., the label of oil film $\mathbf{L}_o$ and the label of seawater $\mathbf{L}_s$. It should be mentioned that if the scene has other objects, such as boats and islands, we can first extract the water area, and then perform the proposed method. In this study, an efficient $k$-means algorithm is performed on the probability map $\mathbf{p}$ to generate the training samples $\mathbf{T}_r=\{\mathbf{L}_o, \mathbf{L}_s\}$.
\subsection{Oil spill detection and optimization}
Given training samples $\mathbf{T}_r=\{\mathbf{L}_o, \mathbf{L}_s\}$ and the dimension-reduced data $\mathbf{Y}$, a widely used spectral classifier, i.e., SVM, is employed to produce the initial detection map $\mathbf{O}$, in which 1\% of the whole training samples are randomly selected from $\mathbf{T}_r$ instead of using all available samples to avoid high computing cost.  Here, the SVM is carried out with LIBSVM library \citep{SVM,FDSI}, which is a Gaussian kernel classification algorithm that allows to process high dimensional data. The parameters in the SVM are decided using five-fold cross validation.\par
Since the SVM is a pixel-wise classifier without considering any spatial information, the probability map obtained by SVM often looks very noisy. Due to the fact that the pixels in oil film and seawater areas are usually highly correlated in the spatial location, this paper introduces the ERW method for refining the initial probability $\mathbf{O}$ to take full advantage of the available spatial information, since the ERW method not only removes the noisy pixels in the detection map but also ensures that the refined probability map can align well with real object boundaries \citep{ERW_kang}.
Specifically, the refined probability maps for oil film and seawater regions are obtained by minimizing the following objective function:
\begin{equation}
\label{eq:erw}
{E^t}({\mathbf{P}_t}) = E_{spatial}^t({\mathbf{P}_t}) + \gamma E_{aspatial}^t({\mathbf{P}_t})
\end{equation}
where $t$ refers to oil film or seawater. The objective function in (\ref{eq:erw}) is comprised of two terms: the spatial and the aspatial terms. The spatial term $\mathbf{E}_{spatial}^t$ can well characterize the spatial correlation among neighboring pixels, which is constructed as follows:
\begin{equation}
E_{spatial}^t({\mathbf{P}_t}) = \mathbf{P}_t^T\mathbf{L}{\mathbf{P}_t}
\end{equation}
where $\mathbf{L}$ is the Laplacian matrix of a weight graph.
\begin{equation}
{L_{ij}} = \left\{ \begin{array}{l}
\sum {{e^{ - \beta {{({v_i} - {v_j})}^2}}}} \quad \;if\;i = j\\
 - {e^{ - \beta {{({v_i} - {v_j})}^2}}}\quad if\;i\;and\;j\;are\;adjacient\;pixels\\
0\quad \quad \quad \quad \quad \quad \;otherwise
\end{array} \right.
\end{equation}
where $v_i$ is the $i$th pixel value and $\beta$ is the parameter. In this work, based on the analysis in the previous publication \citep{ERW_kang}, the parameters $\beta$ and $\gamma$ in the ERW method are set as $\gamma = 10^{-5}$ and $\beta= 710$ for all the experiments.
The second term $E_{aspatial}^t$ incorporates the initial probability maps $\mathbf{O}$ obtained by SVM, which is shown as follows:
\begin{equation}
E_{aspatial}^t({\mathbf{P}_t}) = \sum\limits_{q = 1,q \ne t} {\mathbf{P}_q^T{\mathbf{\Lambda} _q}{\mathbf{P}_q} + {{({\mathbf{P}_t} - 1)}^T}{\mathbf{\Lambda} _t}({\mathbf{P}_t} - 1)}
\end{equation}
where $\mathbf{\Lambda}_t$ represents a diagonal matrix, in which the diagonal elements are composed of the initial probability maps $\mathbf{O}$. Finally, the detection map $\mathbf{\hat C}$ is calculated by choosing the maximum of the refined probability maps $\mathbf{P}_t$:
\begin{equation}
{\mathbf{\hat C}} = \mathop {argmax}\limits_t \;{\mathbf{P}_{t}}
\end{equation}
where $\mathbf{\hat C}$ is the final detection result. According to the method descriptions presented above, Algorithm \ref{alg:Framework} summarizes the pseudocode of the proposed method.

\begin{algorithm}[!tp]
\caption{oil spill detection method}
\label{alg:Framework}
\renewcommand{\algorithmicrequire}{\textbf{Input:}}
\renewcommand{\algorithmicensure}{\textbf{Output:}}
\begin{algorithmic}[1]
\REQUIRE ~~\\
		Input hyperspectral image $\mathbf{I}$ ;\\
\ENSURE ~~\\
    Detection map $\mathbf{\hat C}$\\
\STATE \textbf{Begin}
\STATE According to (\ref{eq:denoising}), remove the severe noisy bands from the input $\mathbf{I}$, so as to obtain noise removal data $\mathbf{S}$.
\STATE According to (\ref{eq:kpca}), reduce the spectral dimension of $\mathbf{S}$, so as to obtain the dimension-reduced image $\mathbf{Y}$.
\STATE Transform dimension reduced image $\mathbf{Y}$ into two-dimensional data, i.e., $\mathbf{\hat{Y}}\in \mathbb{R}^{D\times N}$.
\STATE According to (\ref{eq:prob}), calculate the detection probability $\mathbf{p}_i$ of each pixel $i$.
\STATE Construct the training set $\mathbf{T}_r=\{\mathbf{L}_o, \mathbf{L}_s\}$ by performing the $k$-means algorithm on the probability map $\mathbf{p}$.
\STATE Classify the hyperspectral image $\mathbf{I}$ with the SVM classifier, so as to obtain the initial detection map $\mathbf{O}$.
\STATE According to (\ref{eq:erw}), optimize the initial detection map $\mathbf{O}$ so as to obtain the final detection map $\mathbf{\hat C}$.
\STATE \textbf{Return} $\mathbf{\hat C}$
\STATE \textbf{End}
\end{algorithmic}
\end{algorithm}

\section{Data sets}
\label{sec:data}
\subsection{Study area}
The study area is situated in the Gulf of Mexico, North American continent, around $25^{\circ}$ N $90^{\circ}$ W. As the biggest marine oil disaster in the US history on April 20, 2010, more than 200 million gallons of crude oil are released into the Gulf of Mexico. To better monitor and clean up the oil spill, it is necessary to detect the oil spill region. Hyperspectral data, which provides rich spectral information from the visible to the infrared spectrum, are a good candidate for oil spill detection.
\subsection{HOSD database}
 According to the study area, we create a large-scale hyperspectral database with the AVIRIS sensor from different test sites. The constructed database, called as Hyperspectral Oil Spill Database (HOSD), is the first public oil spill detection dataset, which can act as a data source to develop state-of-the-art approaches for oil spill detection. Compared with oil spill datasets used in other publications, HOSD has the advantages of wide distribution, large coverage, and it also contains the largest number of data. The reference maps of all studied sample images are manually annotated by using the Environment for Visualizing Images (ENVI) software. The oil spill areas are labeled pixel by pixel under the guidance of the field experts. These datasets have been processed with an atmospheric correction model in ENVI 5.3 software before oil spill detection, where the Fast Line-of-Sight Atmospheric Analysis of Spectral Hypercube (FLAASH) is adopted. Table \ref{tab:hosd} lists some features of the HOSD. The spectral coverage is from 365 nm to 2500 nm. Owing to different altitudes in different routes, the spatial resolution of the images is also different. The HOSD will be made freely available\footnote{https://github.com/PuhongDuan/HOSD} for research purposes.
\begin{table}[htbp]
  \centering
  \caption{Some features of the HOSD.}
    \begin{tabular}{cccc}\toprule
    Data  & \multicolumn{1}{l}{Spatial size} & \multicolumn{1}{l}{Resolution} & \multicolumn{1}{l}{Fight time} \\\midrule
    GM1   & 1200*633 & 7.6m  & 5/17/2010 \\
    GM2   & 1881*693 & 7.6m  & 5/17/2010 \\
    GM3   & 1430*691 & 7.6m  & 5/17/2010 \\
    GM4   & 1700*691 & 7.6m  & 5/17/2010 \\
    GM5   & 2042*673 & 7.6m  & 5/17/2010 \\
    GM6   & 2128*689 & 8.1m  & 5/18/2010 \\
    GM7   & 2302*479 & 3.3m  & 7/09/2010 \\
    GM8   & 1668*550 & 3.3m  & 7/09/2010 \\
    GM9   & 1643*447 & 3.2m  & 7/09/2010 \\
    GM10  & 1110*675 & 7.6m  & 5/17/2010 \\
    GM11  & 1206*675 & 7.6m  & 5/17/2010 \\
    GM12  & 869*649 & 7.6m  & 5/06/2010 \\
    GM13  & 1135*527 & 3.2m  & 7/09/2010 \\
    GM14  & 1790*527 & 3.2m  & 7/09/2010 \\
    GM15  & 1777*510 & 3.3m  & 7/09/2010 \\
    GM16  & 1159*388 & 3.2m  & 7/09/2010 \\
    GM17  & 1136*660 & 7.6m  & 5/17/2010\\
    GM18  & 1047*550  & 3.3m  & 7/09/2010\\\bottomrule
    \end{tabular}%
  \label{tab:hosd}%
\end{table}%

\begin{figure}[!tp]
\footnotesize
\setlength{\abovecaptionskip}{0.cm}
\setlength{\belowcaptionskip}{-0.cm}
\centering
\centerline{\subfigure[AUC]{\includegraphics[scale=0.3]{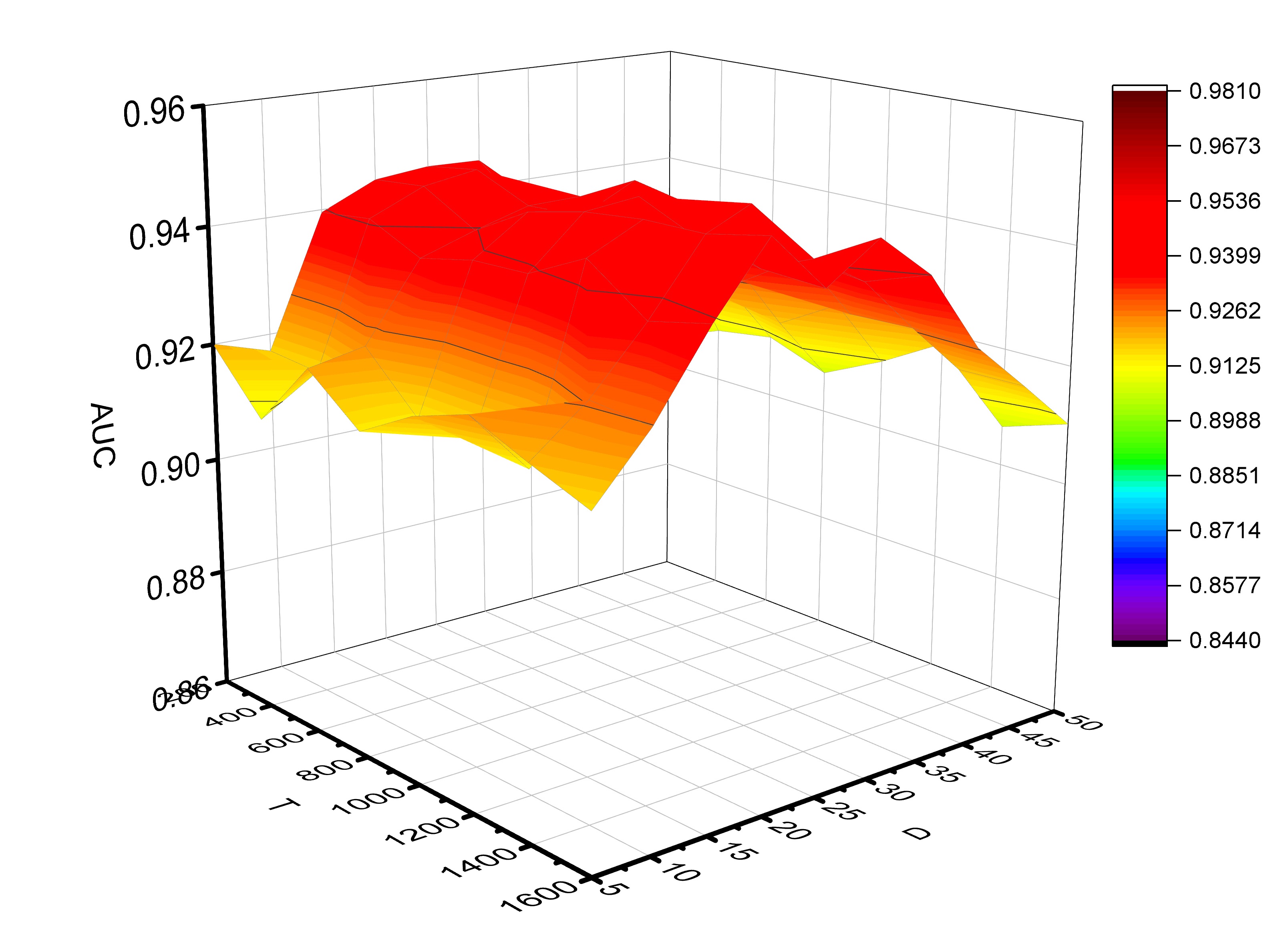}}
			\subfigure[DP]{\includegraphics[scale=0.3]{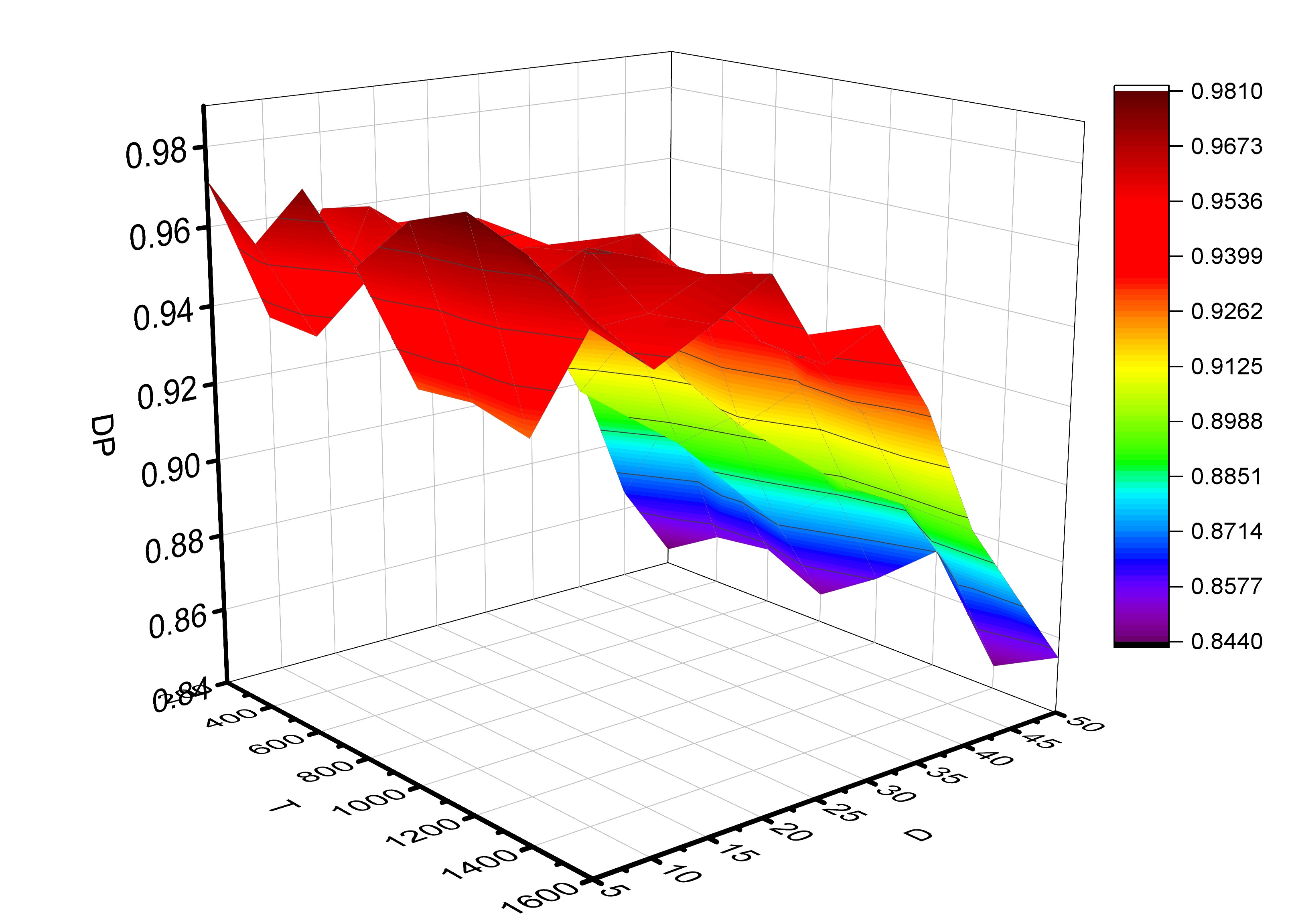}}}
\caption{The sensitivity analysis of the two parameters $D$ and $T$.}
\label{pa}
\end{figure}

\section{Experiments}
\label{sec:exp}

To examine the effectiveness of the proposed method for oil spill detection, several state-of-the-art classification and detection methods are adopted for comparison, including a high dimensional data clustering method, unsupervised classification methods, object detection methods, and a supervised oil spill detection method. For the high dimension data clustering method, a scalable exemplar-based subspace clustering (SESC) method \citep{SERC} was adopted as a competitor, since it can obtain satisfactory clustering performance for high dimensional and class-imbalanced data compared to other clustering methods. For the case of a hyperspectral unsupervised classification method, a representative unsupervised method derived from rank-two nonnegative matrix factorization (R2NMF) \citep{H2NMF} was used as the comparison algorithm. For the case of the hyperspectral unsupervised object detection methods, a popular object detection approach based on low-rank and sparse matrix decomposition (LRSMD) \citep{LRSMD} was employed to achieve the oil spill detection. Another recently proposed object detection method based on kernel isolation forest (KIF) \citep{KIF} was considered for comparison purposes because it shows state-of-the-art performance in object detection. For the supervised oil spill detection method, a PCA-based minimum distance (PCAMD) detection method \citep{PCAMD} was used as for comparison, in which 25 principal components were preserved and the number of training samples was 1\% for oil film and seawater. The implementation of these approaches was achieved by using the publicly available codes. For all parameters of these approaches, we mainly comply with the recommendations of the authors in the related publications.\par
To objectively evaluate the detection performance of different approaches, two widely used objective indexes, i.e., area under curve (AUC) \citep{FAWCETT2006861} and detection precision (DP) \citep{WANG2021107741}, for object detection are adopted. \par
1) AUC: Given a detection result and a reference map, the AUC is defined as:
\begin{equation}
AUC = \int_{ - \infty }^{ + \infty } {TPR(H)FP{R^{'}}(H)dH}
\end{equation}
where the true positive rate $TPR(H)$ measures how many true positive samples happen among all positive results, when the threshold for the detection result is $H$. $FPR$ represents how many false positive samples happen among all negative results. The main superiority of the AUC index is that it relies only on the order of the pixels rather than absolute detection values.\par
2) DP: DP calculates the percentage of the true positives to the total number of positive samples, which is defined as
\begin{equation}
DP = \frac{TP}{TP+FP}
\end{equation}
where TP and FP denote true positive and false positive, respectively. The higher the value of
this index, the better the oil spill detection performance is.

\begin{figure*}[!tp]
\centering
\includegraphics[width=6.6in,height=8.6in]{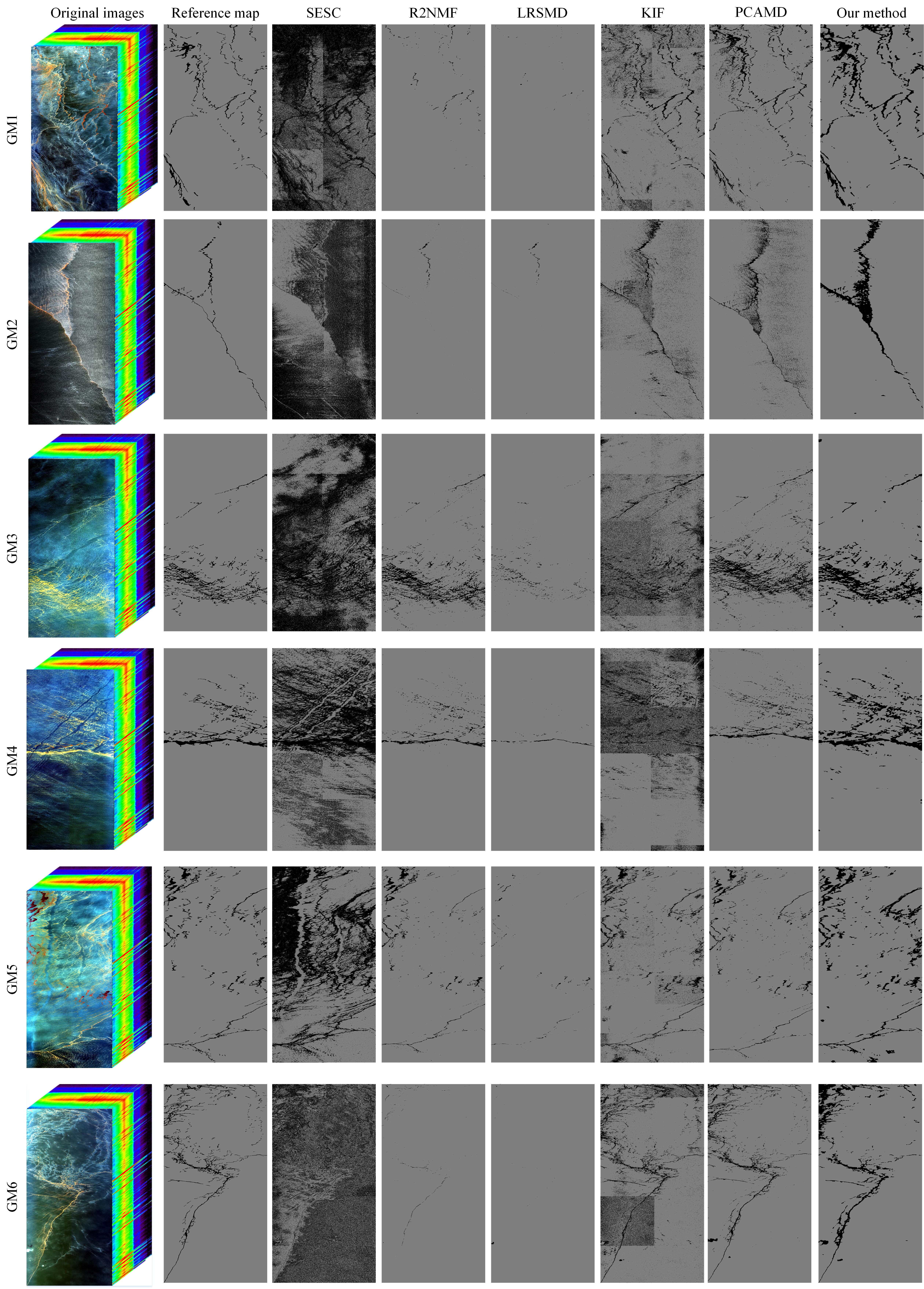}
\caption{ The original images and the detection results of all considered methods on GM1 to GM6 images. From left to right: original images, reference maps, and detection results of different approaches.}
\label{re1}
\end{figure*}
\begin{figure*}[!tp]
\centering
\includegraphics[width=6.6in,height=8.6in]{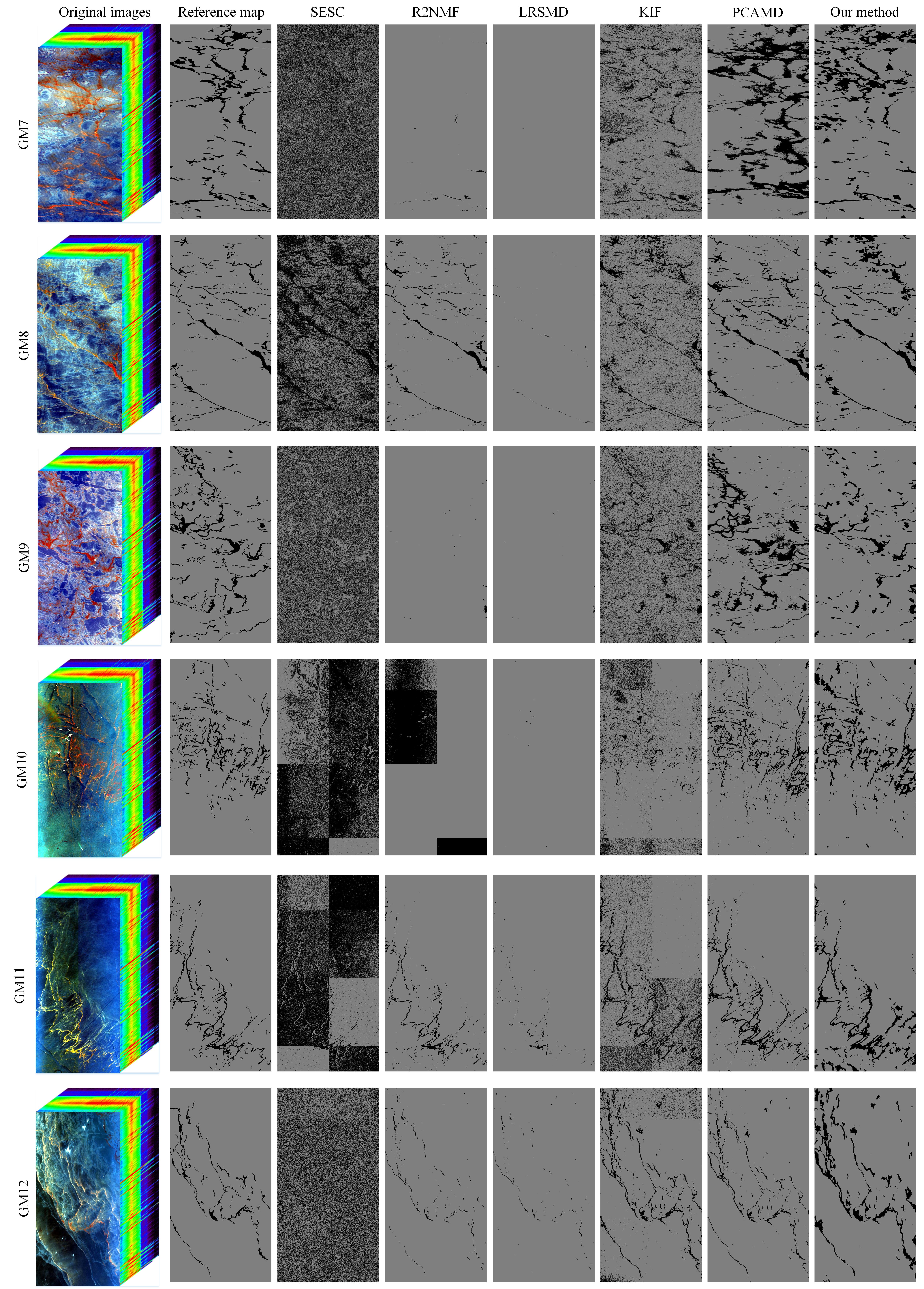}
\caption{ The original images and the detection results of all considered methods on GM7 to GM12 images. From left to right: original images, reference maps, and detection results of different approaches.}
\label{re2}
\end{figure*}

\begin{figure*}[!tp]
\centering
\includegraphics[width=6.6in,height=8.6in]{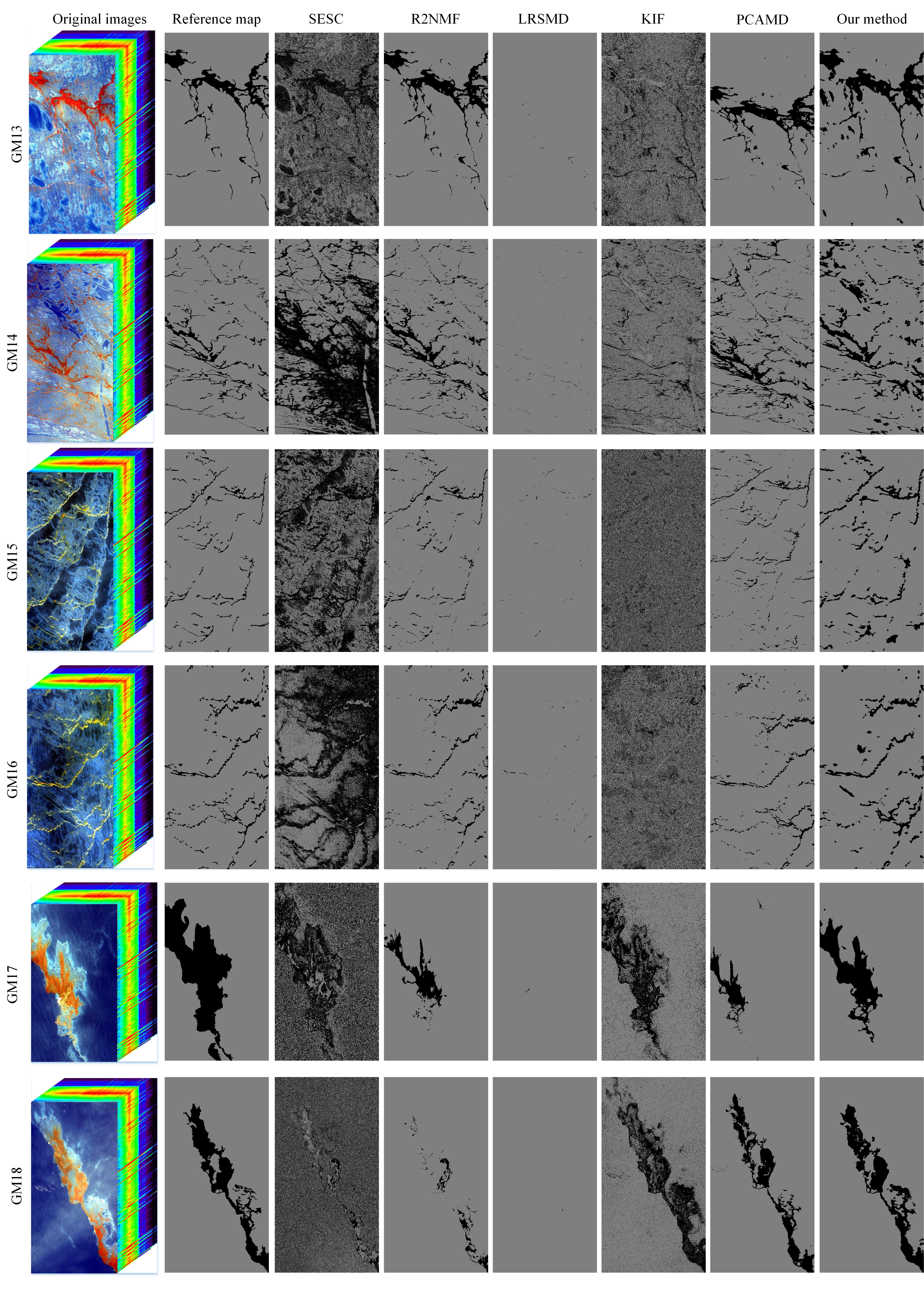}
\caption{ The original images and the detection results of all considered methods on GM13 to GM18 images. From left to right: original images, reference maps, and detection results of different approaches.}
\label{re3}
\end{figure*}
\subsection{Parameter analysis}
In this work, there are two parameters that need to be determined, i.e., the number of the fused data $D$ and the number of isolation trees $T$. The sensitivity analysis of the two parameters is conduced on the GM1 image. Parameter $D$ varies from 5 to 50 with the step size of 5, and parameter $T$ ranges from 200 to 1600 with the step size of 200. Fig. \ref{pa} presents the AUC and DP values of the proposed method with different parameters. From Fig. \ref{pa}, it is found out that the proposed method yields the highest AUC value when $D$ and $T$ are set as 25 and 800, respectively. In addition, with this parameter setting, the DP value of the proposed method is also relatively high. Based on this observation, $D=25$ and $T=800$ are considered as the default parameters in this paper. The following experiments also illustrate that the proposed method indeed produces promising detection results with this parameter setting.

\subsection{Detection results}
\sloppy{}
The original images and detection results of different methods are presented in Figs. \ref{re1}-\ref{re3}. By comparing the detection results, it can be observed that the object detection methods based on LRSMD and KIF cannot work well for oil spill detection. The LRSMD method can only detect a small amount of oil spill regions, since the Mahalanobis distance cannot well model the spectral difference between the seawater and oil film. By contrast, The KIF method exhibits a little improvement in terms of detection performance. However, this method produces very \lq\lq noisy\rq\rq\ resulting maps, and some images can hardly be identified, such as GM13, GM14, GM15, and GM16.\par

For the high dimensional clustering method, the SESC method produces similar results with the KIF method. The oil film and seawater cannot be distinguished by spectral clustering. This is due to the low discrimination between seawater and oil film, and thus, it is difficult to effectively detect the oil spill regions from similar spectrum. For example, most seawater regions in the detection maps are misclassified into oil spill regions. Compared to the clustering-based method, the unsupervised R2NMF classification method presents better detection performance on several sample images (e.g., GM13-GM16). The reason is that the hierarchical clustering framework is used to iteratively optimize the detection results. However, the R2NMF method cannot obtain satisfactory performance on some complicated scenes, such as GM1, GM2, GM9, and GM10.\par
For the supervised oil spill detection method, it is easy to observe that the PCAMD method outperforms other unsupervised approaches in terms of detection performance since it takes advantage of the training samples of the reference map. However, due to the inner-class differences of oil spills, the oil spills in the edge regions cannot be well identified using the 1\% of the training samples. In this case, some pixels belong to the oil spill class are misclassified as seawater, such as the bottom left corner in the GM1 image. \par
 As can be seen, the proposed method can well identify the oil spill regions in all sample images in comparison to other approaches. Moreover, the shapes of the oil spill regions are also well aligned with the reference maps. The satisfactory detection results obtained by the proposed method are mainly due to several reasons: First, the noisy band removal can effectively alleviate the interference of severe noisy bands to detection performance. This step is able to greatly boost the robustness of the proposed method. Second, the isolation forest method produces relatively true training samples, even though it does not involve human annotation. Third, the ERW-based optimization process takes full advantage of the spatial correlations among neighboring pixels. This optimization is beneficial for improving the detection accuracy.\par

Tables \ref{tab:auc} and \ref{tab:DP} give the objective results of the considered approaches on the HOSD database. It is obvious that the proposed method obtains competitive performance among all considered methods in terms of AUC and DP, which is consistent with the visual effects. The PCAMD method is better than the proposed method on five sample images. The reason is that the PCAMD is a supervised approach, in which the training set is directly chosen from the reference label. The preparation of training samples is time-demanding and requires a large amount of manpower. On the contrary, the proposed method aims to automatically obtain the pseudo-labeled training samples, which is of high interest in real applications. Generally, it can be observed from Tables \ref{tab:auc} and \ref{tab:DP} that our method produces the highest detection accuracy compared to other approaches on the HOSD database.

\begin{table}[!tp]
  \centering
  \caption{The AUC results obtained by all considered approaches on the HOSD.}
    \begin{tabular}{ccccccc}\toprule
    AUC   & SESC  & R2NMF & LRSMD & KIF   & PCAMD & Our method \\\midrule
    GM01  & 0.6848  & 0.5578  & 0.5073  & 0.8109  & 0.7787  & \textbf{0.9472} \\
    GM02  & 0.4221  & 0.5747  & 0.5515  & 0.8632  & 0.8443  & \textbf{0.9523} \\
    GM03  & 0.7196  & 0.9395  & 0.5809  & 0.7720  & \textbf{0.9435} & 0.9428  \\
    GM04  & 0.8107  & 0.7505  & 0.5407  & 0.7249  & 0.8754  & \textbf{0.9655} \\
    GM05  & 0.8201  & 0.7734  & 0.5413  & 0.8483  & 0.8296  & \textbf{0.9185} \\
    GM06  & 0.4949  & 0.5258  & 0.5014  & 0.8534  & 0.8556  & \textbf{0.8632} \\
    GM07  & 0.5608  & 0.5131  & 0.4999  & 0.6966  & \textbf{0.8907} & 0.8452  \\
    GM08  & 0.7175  & 0.8279  & 0.5085  & 0.7852  & \textbf{0.9311} & 0.8956  \\
    GM09  & 0.3954  & 0.5001  & 0.4997  & 0.7223  & \textbf{0.8825} & 0.7914  \\
    GM10  & 0.3212  & 0.5928  & 0.4999  & 0.6652  & 0.7487  & \textbf{0.9022} \\
    GM11  & 0.6014  & 0.7732  & 0.5521  & 0.8642  & 0.8321  & \textbf{0.9303} \\
    GM12  & 0.4827  & 0.6385  & 0.5605  & 0.8953  & 0.8457  & \textbf{0.9352} \\
    GM13  & 0.7304  & \textbf{0.9671}  & 0.5041  & 0.5350  & 0.9649 & 0.9416  \\
    GM14  & 0.7834  & 0.9310  & 0.5179  & 0.6601  & \textbf{0.9461} & 0.8785  \\
    GM15  & 0.7492  & 0.8372  & 0.5227  & 0.4983  & 0.8939  & \textbf{0.9094} \\
    GM16  & 0.5853  & 0.8461  & 0.5187  & 0.4815  & \textbf{0.9210} & 0.8894  \\
    GM17  & 0.6102  & 0.6263  & 0.5003  & \textbf{0.7955} & 0.7143  & 0.7563  \\
    GM18  & 0.5158  & 0.5951  & 0.4999  & 0.8298  & 0.8485  & \textbf{0.9469} \\\midrule
    Mean    & 0.6114  & 0.7095  & 0.5226  & 0.7390  & 0.8637  & \textbf{0.9006} \\\bottomrule
    \end{tabular}%
  \label{tab:auc}%
\end{table}

\begin{table}[htbp]
  \centering
  \caption{The DP results obtained by all considered approaches on the HOSD.}
    \begin{tabular}{ccccccc}\toprule
    DP    & SESC  & R2NMF & LRSMD & KIF   & PCAMD & Our method \\\midrule
    GM01  & 0.9659  & 0.1161  & 0.0146  & 0.7080  & 0.5721  & \textbf{0.9615} \\
    GM02  & 0.3941  & 0.1497  & 0.1030  & 0.7953  & 0.7453  & \textbf{0.9654} \\
    GM03  & \textbf{0.9858} & 0.9032  & 0.1624  & 0.7458  & 0.9185  & 0.9290  \\
    GM04  & \textbf{0.9973} & 0.5021  & 0.0815  & 0.6854  & 0.7572  & 0.9721  \\
    GM05  & \textbf{0.9856} & 0.5499  & 0.0826  & 0.7375  & 0.6652  & 0.8815  \\
    GM06  & 0.4679  & 0.0517  & 0.0029  & 0.8046  & 0.7171  & \textbf{0.7626} \\
    GM07  & 0.6227  & 0.0262  & 0.0000  & 0.5148  & \textbf{0.9422} & 0.7600  \\
    GM08  & \textbf{0.9639} & 0.6597  & 0.0169  & 0.7003  & 0.8791  & 0.8535  \\
    GM09  & 0.2427  & 0.0014  & 0.0008  & 0.6241  & \textbf{0.8807} & 0.6051  \\
    GM10  & 0.2758  & 0.4524  & 0.0000  & 0.4181  & 0.7362  & \textbf{0.8526} \\
    GM11  & 0.7813  & 0.5477  & 0.1043  & 0.8606  & 0.6656  & \textbf{0.9600} \\
    GM12  & 0.4244  & 0.2777  & 0.1216  & 0.8308  & 0.6985  & \textbf{0.9224} \\
    GM13  & 0.8783  & 0.9576  & 0.0081  & 0.3846  & \textbf{0.9583} & 0.9404  \\
    GM14  & \textbf{0.9901} & 0.8971  & 0.0363  & 0.5166  & 0.9558  & 0.8341  \\
    GM15  & \textbf{0.9702} & 0.6818  & 0.0455  & 0.3942  & 0.7997  & 0.8712  \\
    GM16  & 0.7081  & 0.6993  & 0.0374  & 0.3253  & 0.8549 & \textbf{0.8944}  \\
    GM17  & \textbf{0.7112} & 0.2526  & 0.0006  & 0.6793  & 0.4289  & 0.5148  \\
    GM18  & 0.5026  & 0.1904  & 0.0000  & 0.7973  & 0.6984  & \textbf{0.9115} \\\midrule
    Mean  & 0.7149  & 0.4398  & 0.0455  & 0.6401  & 0.7708  & \textbf{0.8551} \\\bottomrule
    \end{tabular}%
  \label{tab:DP}%
\end{table}%
\section{Discussion}
\label{sec:dis}
\subsection{The influence of the bands corrupted by severe noise}
\begin{figure}[htbp]
\footnotesize
\setlength{\abovecaptionskip}{0.cm}
\setlength{\belowcaptionskip}{-0.cm}
\centering
\centerline{\subfigure[AUC]{\includegraphics[scale=0.36]{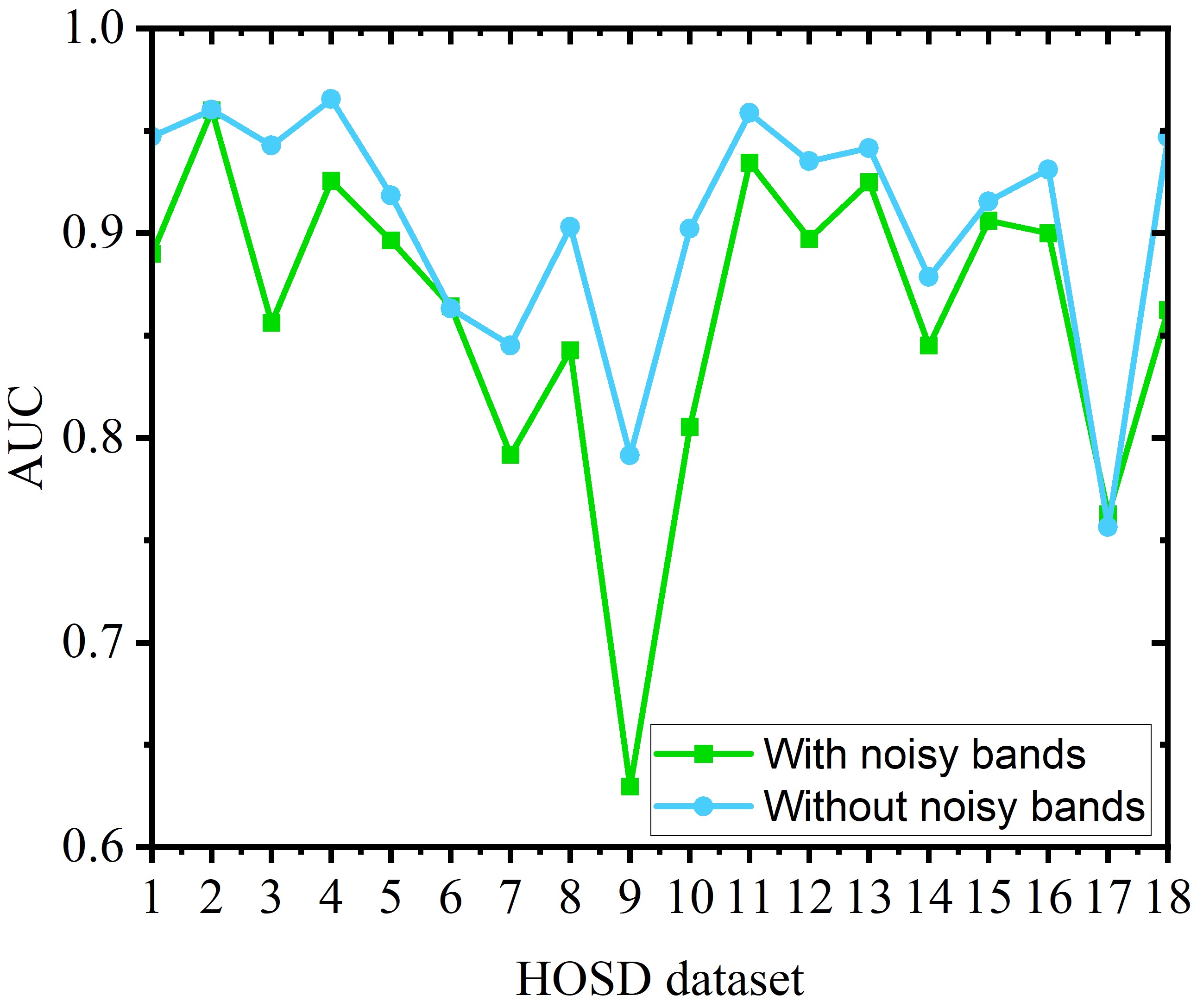}}
			\subfigure[DP]{\includegraphics[scale=0.36]{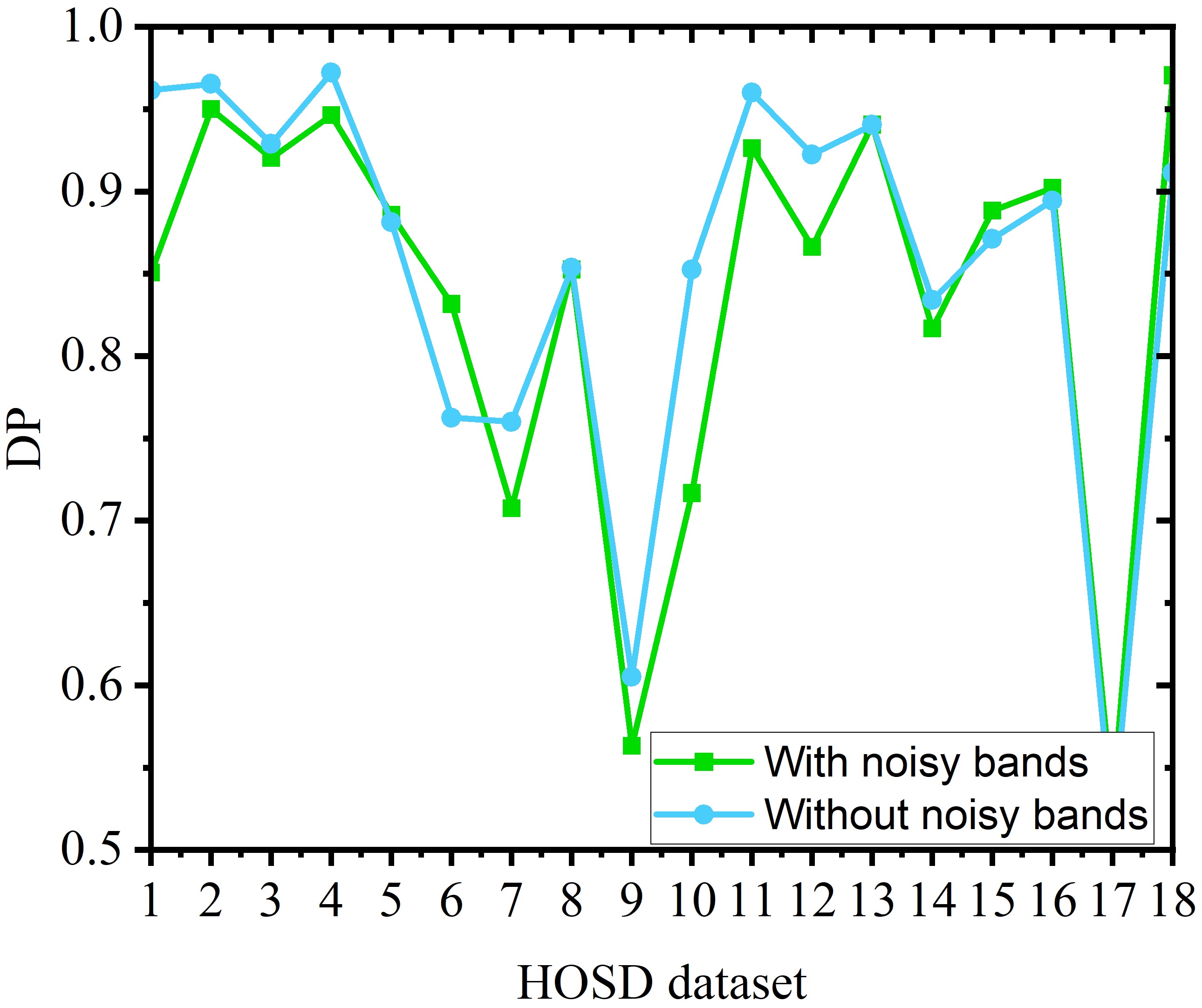}}}
\caption{The detection accuracies of the proposed approach with or without the bands corrupted by severe noise. }
\label{fig:noise}
\end{figure}

In this subsection, we discuss the influence of the bands corrupted by severe noise on the detection performance of the proposed method.
An experiment is conducted on the HOSD database with or without the bands corrupted by serious noise. The detection accuracy is shown in Fig. \ref{fig:noise}. Through comparing the experimental results, it is found that when the corrupted bands are removed, the detection accuracies obtained by the proposed method tend to increase. This experiment verifies the effectiveness of the Gaussian statistical method in automatically removing the bands corrupted by serious noise. In addition, it also exhibits the robustness of our method.

\begin{figure}[htbp]
\footnotesize
\setlength{\abovecaptionskip}{0.cm}
\setlength{\belowcaptionskip}{-0.cm}
\centering
\centerline{\subfigure[AUC=0.9166]{\includegraphics[scale=0.16]{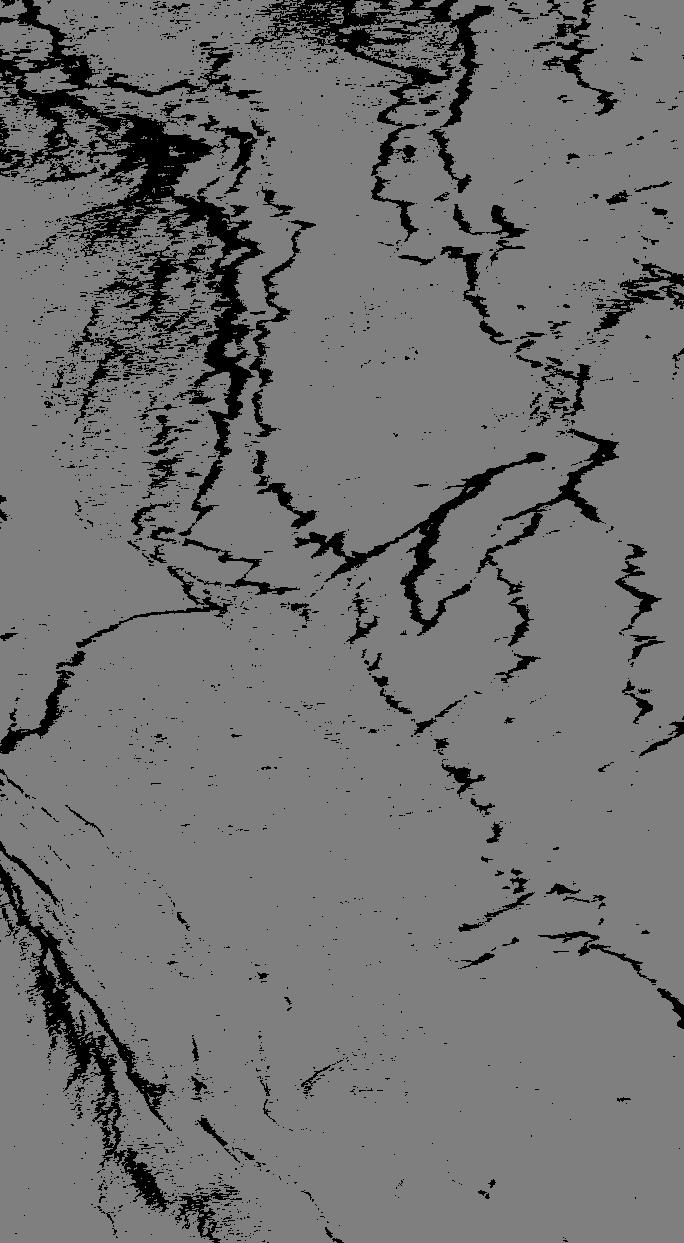}}
			\subfigure[AUC=0.9472]{\includegraphics[scale=0.16]{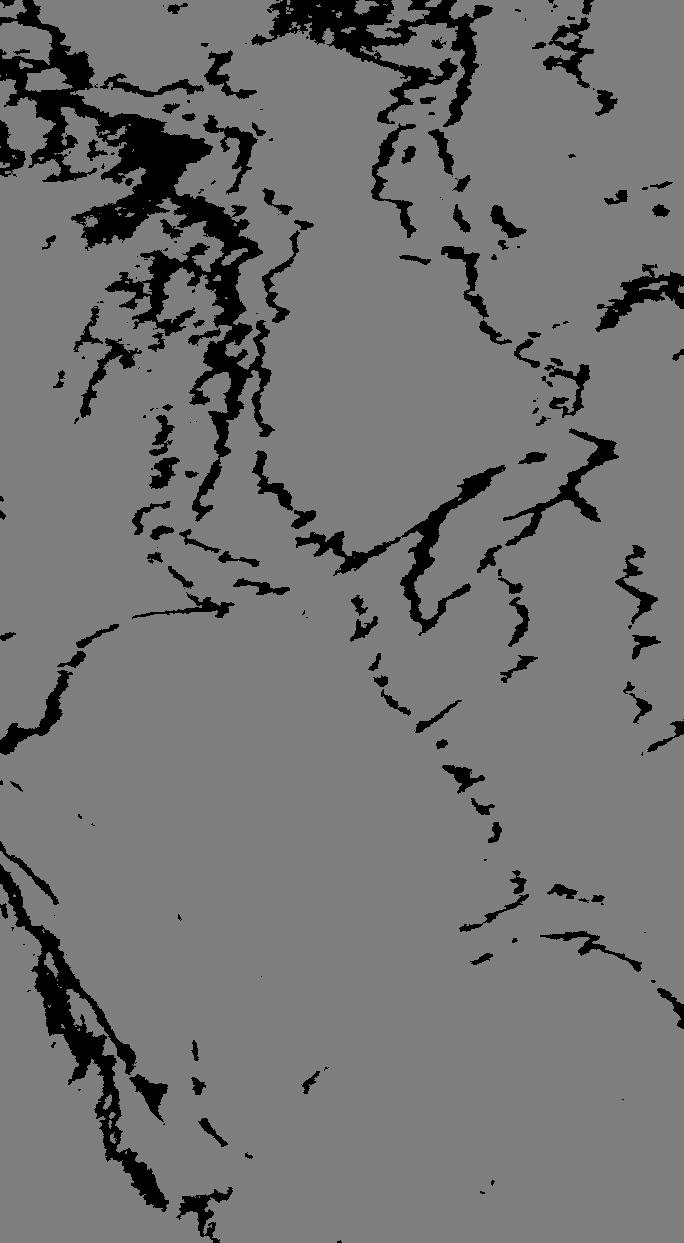}}}
\caption{The influence of the ERW optimization. (a) The proposed method without the ERW step. (b) The proposed method.}
\label{fig:ERW}
\end{figure}
\subsection{The influence of the ERW optimization}
Due to the spectral similarity between different objects, the classification map obtained by the spectral classifier (e.g., SVM) usually contains the salt-and-pepper noise. The spatial information in HSIs can play a complementary role in reducing misclassification. This is mainly because of the fact that the pixels belonging to the same object are often adjacent in the spatial location. Here, we discuss the influence of the ERW optimization to oil spill detection accuracy. An experiment is conducted on the GM1 image. Fig. \ref{fig:ERW} gives the detection maps obtained by the proposed method with or without the optimization step. It is easy to observe that the proposed method without ERW optimization tends to yield a noisy detection map (see Fig. \ref{fig:ERW}(a)). When the spatial information is considered by using ERW, the misclassified pixels in homogeneous regions can be effectively smoothed out (see Fig. \ref{fig:ERW}(b)). In addition, the detection accuracies are increased by AUC=3.06\% and DP=5.37\%, respectively. Therefore, the ERW-based optimization is an effective scheme to increase the detection performance in terms of visual quality and objective index, which can ensure that adjacent pixels belong to the same object in the final detection result.

\section{Conclusions}
\label{sec:con}
In this study, we developed an unsupervised oil spill detection approach for hyperspectral images. The proposed approach, which is based on the robust isolation forest, demonstrates several attractive characteristics:
1) Hyperspectral data are often contaminated by noise, which downgrades the quality of the detection step. This paper introduces a Gaussian statistical method to automatically remove the bands contaminated by severe noise. Therefore, the proposed method is robust to corrupted bands.
2) To decrease the involvement of human labor, this paper proposes an automatic method based on the isolation forest to select training samples of the input data. Accordingly, it does not require manual labeling, which can be better applied in real applications.
3) This paper makes full use of the spatial information among adjacent pixels, and thus, the \lq\lq noise \rq\rq\ in the detection maps can be greatly reduced and the detection performance is also improved.\par
In addition, we constructed a large-scale oil spill detection dataset, i.e., HOSD, with few images. The HOSD is a freely-available dataset and can be used to achieve oil spill detection or train deep models. We have tested the performance of the proposed approach on the HOSD. Experimental results have confirmed that the proposed approach can outperform other advanced approaches including high dimensional data clustering, unsupervised object detection methods, and supervised oil spill detection technique.\par
Despite the promising performance of the proposed method, we have to admit that there is still room for improvement. In the proposed method, although we can generate automatically the training samples, it is difficult to guarantee that the generated samples are highly accurate. In consequence, an effective scheme will be developed to generate a high-quality training set in the future.
\section*{Acknowledgment}
{We would like to thank Junfang Yang and Yi Ma from the First Institute of Oceanography, State Oceanic Administration, China, for providing help in the labeling oil spill regions, and NASA for providing AVIRIS images. This paper is supported by the National Key R \& D Program of China (Grant No. 2021YFA0715203), the Major Program of the National Natural Science Foundation of China (Grant No. 61890962), the National Natural Science Foundation of China (Grant No. 61871179 and 62201207), the Scientific Research Project of Hunan Education Department (Grant No. 19B105), the National Science Foundation of Hunan Province (Grant No. 2019JJ50036 and 2020GK2038), the Hunan Provincial Natural Science Foundation for Distinguished Young Scholars (Grant No. 2021JJ022), the Huxiang Young Talents Science and Technology Innovation Program (2020RC3013), and the Changsha Natural Science Foundation (Grant No. kq2202171).}

\bibliography{refs}

\end{document}